\documentclass[conference]{IEEEtran}
\usepackage[utf8]{inputenc}
\usepackage{comment}

\usepackage[detect-all]{siunitx}

\usepackage{multirow}

\usepackage{graphicx}
\graphicspath{ {./images/} }
\usepackage{subfigure}
\usepackage{soul}

\IEEEoverridecommandlockouts

\begin{document}

\title{Levels of Automation for a Mobile Robot Teleoperated by a Caregiver}

\author{
\IEEEauthorblockN{Samuel Olatunji\IEEEauthorrefmark{1}}
\IEEEauthorblockA{\textit{Department of Industrial Engineering and Management} \\
\textit{Ben-Gurion University of the Negev}\\
Be’er Sheva, Israel \\
olatunji@post.bgu.ac.il}
\and
\IEEEauthorblockN{Andre Potenza\IEEEauthorrefmark{1}}
\IEEEauthorblockA{\textit{Center for Applied Autonomous Sensor Systems (AASS)} \\
\textit{Örebro University}\\
Örebro, Sweden \\
andre.potenza@oru.se}
\and
\IEEEauthorblockN{Andrey Kiselev}
\IEEEauthorblockA{\textit{Center for Applied Autonomous Sensor Systems (AASS)} \\
\textit{Örebro University}\\
Örebro, Sweden \\
andrey.kiselev@oru.se}
\and
\IEEEauthorblockN{Tal Oron-Gilad}
\IEEEauthorblockA{\textit{Department of Industrial Engineering and Management} \\
\textit{Ben-Gurion University of the Negev}\\
Be’er Sheva, Israel \\
orontal@bgu.ac.il}
\and
\IEEEauthorblockN{Amy Loutfi}
\IEEEauthorblockA{\textit{Center for Applied Autonomous Sensor Systems (AASS)} \\
\textit{Örebro University}\\
Örebro, Sweden \\
amy.loutfi@oru.se}
\and
\IEEEauthorblockN{Yael Edan}
\IEEEauthorblockA{\textit{Department of Industrial Engineering and Management} \\
\textit{Ben-Gurion University of the Negev}\\
Be’er Sheva, Israel \\
yael@bgu.ac.il}
\thanks{$^{*}$Equal contribution}
}

\maketitle

\begin{abstract}
Caregivers in eldercare can benefit from telepresence robots that allow them to perform a variety of tasks remotely. In order for such robots to be operated effectively and efficiently by non-technical users, it is important to examine if and how the robotic system's level of automation (LOA) impacts their performance.  
The objective of this work was to develop suitable LOA modes for a mobile robotic telepresence (MRP) system for eldercare and assess their influence on users' performance, workload, awareness of the environment and usability at two different levels of task complexity.
For this purpose, two LOA modes were implemented on the MRP platform: assisted teleoperation (low LOA mode) and autonomous navigation (high LOA mode). The system was evaluated in a user study with 20 participants, who, in the role of the caregiver, navigated the robot through a home-like environment to perform control and perception tasks.
Results revealed that performance improved in the high LOA when task complexity was low. However, when task complexity increased, lower LOA improved performance. This opposite trend was also observed in the results for workload and situation awareness. We discuss the results in terms of the LOAs' impact on users' attitude towards automation and implications on usability.
\end{abstract}


\section{Introduction}
Many older adults prefer aging in place within their own homes, which has been associated with increased well-being~\cite{rueangsirarak2012fall}. 
Moreover, as the ratio between the aging population and caregivers continually increases~\cite{redfoot2013aging} and families live farther apart, this circumstance often becomes a necessity. 
Although many older adults manage to remain largely independent, they may require occasional help with activities of daily living from a caregiver~\cite{abou2020systematic}. One way to facilitate this is through the use of mobile robotic telepresence (MRP) systems~\cite{kristoffersson2013review}. MRP refers to the activity of remote controlling a robot that is able to move through its environment, thereby enabling its user (referred to as remote user or operator) to interact with other people (local users) within that same physical space~\cite{kristoffersson2013review}. As robotic hardware keeps evolving and sophisticated sensors and actuators become affordable, users will be enabled to perform more complex tasks -- for example, as in the present use case, to support older adults in their daily lives~\cite{olatunji2020user} or when social distancing is required \cite{greenstone2020does}. In the future, caregivers who cannot be physically present in the older adult's residence will be able to use telepresence robots to visit. These systems afford a high level of independence and wide range of action, as seen in several projects focusing on the development of telepresence robots to support older adults in a range of daily activities~\cite{zafrani2019towards}.
Examples of previous work include developing a research platform (GiraffPlus~\cite{Coradeschi2013}), design recommendations and functionalities (ExCITE EU project~\cite{Orlandini2016}), and developing social navigation capabilities (the TERESA project~\cite{shiarlis2015teresa}). 
The present study focuses on two specific constructs, levels of automation (LOA) and task complexity, which were evaluated 
for a teleoperated mobile robot in terms of performance, workload, situation awareness and usability. 

LOA defines the degree to which automation is employed in a given task~\cite{endsley1999level, beer2014toward, olatunji2020LOALOT}. Several short-term evaluations of MRP for telecare in home environments revealed 
maneuvering challenges and high workload when manually driving the robot. Hence, this study focuses on the design of LOA for navigation ~\cite{cesta2012evaluating, kiselev2012using}. If navigation is automated efficiently, it can free the operator to direct their attention to other tasks and subtasks~\cite{parasuraman2008situation}. Furthermore, as tasks become increasingly complex, it may be difficult for an operator to conduct the task sufficiently well manually \cite{crandall2002characterizing}. Therefore, navigation is a critical function for which LOA can be introduced, especially in the context of different levels of task complexity.

Situation awareness (SA) is described as the degree to which an operator, at any point in time while controlling or monitoring a system (e.g., a robot), is aware of its current internal state and the state of the external environment in which it is located and acting~\cite{endsley1995out, endsley2001designing}. In the literature SA is frequently defined as a qualitatively incremental variable spanning three levels: The ability to perceive the incoming and available data that is relevant to the current task or context, understanding the meaning and implications of that data as it pertains to the current state, and finally the ability to make predictions about how the situation is likely to develop in the (near) future if kept on its current trajectory~\cite{endsley2001designing}.
Mental workload is a measure of the degree to which a person's executive cognitive functions are being occupied at a given time~\cite{cain2007review}, for example as a result of focusing on one or several tasks which demand sustained allocation of these processes.

Task complexity has been identified in previous research as a critical factor impacting performance~\cite{crandall2002characterizing} and influencing the LOA design in human-robot interaction~\cite{beer2014toward,honig2018toward}. It depends predominantly on properties of the task (\textit{objective complexity}) and the perception of the human operator (\textit{subjective complexity})~\cite{rasmussen2015task}. Further dimensions of complexity include \textit{component complexity} -- the number of distinct actions that the human operator must execute or number of informational cues that should be processed (e.g., the number and types of subtasks to be managed individually)~\cite{olsen2003metrics}; \textit{coordinative complexity} -- the nature of relationships between task inputs and task products, the strength of these relationships, simultaneous action requirements, as well as the sequencing of inputs (e.g., timing, frequency, intensity and location requirements, level of difficulty)~\cite{campbell1988task}, as well as \textit{dynamic complexity} -- changes in the states of the environment, e.g., cause-effect chains, means-ends connections to which the human operator should adapt, criticality of changes and the degree of human intervention required for these changes~\cite{braarud2001subjective, wood1986task}. 

Only few studies have considered task complexity in the design and evaluation of LOA for MRP systems specifically in the case of non-technical operators (e.g., caregivers)~\cite{steinfeld2006common, michaud2007telepresence, kristoffersson2013review,gutman2021evaluating}. Specific LOA designs suitable and feasible for such MRP scenarios are still lacking~\cite{vagia2016literature}. Previous work~\cite{kiselev2015evaluation} evaluated semi-autonomy features in MRPs designed to relieve the pilot user of some of the mental and physical demand associated with maneuvering the robot. Other studies revealed that adjusting the robot's autonomy in teleoperation tasks can help facilitate their use across a broader range of applications~\cite{kaber2018issues, sheridan1978human}. In the context of a user interacting with a robot, such as in the case of an MRP task, it has been previously argued that dynamic adaptation of control and responsibilities is necessary to accommodate different users and successfully manage a variety of situations and tasks~\cite{potenza2018one}. Autonomy is typically considered as a continuous or discrete spectrum, with direct human control and full autonomy at either end, and any number of intermediate levels in between~\cite{beer2014toward}. However, this view is arguably more appropriate when considered at the task level, since a system may be performing multiple tasks simultaneously at different levels of autonomy. Moreover, most tasks can be decomposed further into subtasks and basic actions, which may be shared between a robot and human operator. 

This work aims to \textbf{develop and evaluate two LOA designs} -- assisted teleoperation (low LOA) mode and autonomous (high LOA) mode -- \textbf{which are functional, suitable and adaptable for users in different task complexity scenarios}. In general, the results of previous LOA experiments did not reveal clear-cut response trends regarding the influence of LOA on operator performance, workload and situation awareness~\cite{kaber2018issues}. 
The outcomes of these previous studies further suggest that some of our intuitions about automation and human performance have been imprecise or incomplete, partly due to LOA evaluations conducted in isolation from other influencing factors and limited number of studies conducted in this area~\cite{onnasch2014human}. In response to the recurring need for more accurate assessments of human performance variables in different contexts, the current research investigates the interaction of the LOA modes with different task complexity levels that may arise in the home environment during teleoperation. Navigating the robot in a more complex task with longer interactions, more waypoints and obstacles can lead to fatigue or loss of SA, on which the MRP system's LOA design may have an impact. User attitude towards the LOA designs at different levels of complexity was investigated in a user study.

Participants took on the role of a caregiver controlling a telepresence robot using different LOA modes -- to navigate through an assisted living apartment, check in on a person and perform a variety of observational tasks. The study comprised four conditions, combining two independent variables in a factorial format: LOA and task complexity.
The objective was to evaluate the influence of the LOA modes on users' performance, workload, SA and usability in a teleoperated task with a focus on navigation.

The next section describes the different aspects of the MRP task in the present study, i.e., the functions to be carried out by the user, their respective allocation in the two LOAs, as well as the process of facilitating the coordination of the user-robot interaction.
Section~\ref{sec:methodology} provides an in-depth description of the study design and methodology, followed by results, discussion and outlook in sections \ref{sec:results}, \ref{sec:discussion} and \ref{sec:conclusions}, respectively.

\section{Function Allocation and LOA Development in MRP}

\label{sec:functions}

The function allocation and LOA development for the MRP were aimed to ensure that tasks (and subtasks) in the system are appropriately allocated to the human, robot, or both, at specific degrees of autonomy while allowing for adaptation as required under varying conditions. Fully manual and fully autonomous modes were not included in the design. Achieving fully autonomous operation is not practical at the moment, as tasks may change and systems are still not developed enough to handle dynamic changes in the home environment. Instead, intermediate levels where functions and tasks are shared between operator and robot are feasible.

The function allocations are based on estimated capacities of the user and robot in a given situation to ensure coordination and collaboration between the human and automation~\cite{dekker2002maba}. 
The functions in the MRP system were identified according to the four-stage 'O-O-D-A loop' information model~\cite{boyd1996essence, endsley1999level, parasuraman2000model}. 
It involves functions related to 1) information acquisition (\emph{Observe}), 
2) information analysis (\emph{Orient}), 3) decision selection (\emph{Decide}) and
4) action implementation (\emph{Act}), for the operation of the robot in the local environment and the remote operator user interface, as shown in Table \ref{tab:1}. 

\begin{table*}[!ht]
\centering
\caption{Functions in the MRP task}
\label{tab:1}
\begin{tabular}{lllll}
\hline
\textbf{System Aspect} & \multicolumn{4}{c}{Functions at different stages}                                                                                                              \\ \cline{2-5} 
                                        & \textbf{Information} & \textbf{Information} & \textbf{Decision}  & \textbf{Action}         \\
                                        & \textbf{Acquisition} & \textbf{Analysis}    & \textbf{Selection} & \textbf{Implementation} \\
                                        & (\textbf{O}bserve)                             & (\textbf{O}rient)                              & (\textbf{D}ecide)                            & (\textbf{A}ct)                                    \\ \hline
Robot                                   & Monitoring the                        & Generating                            & Selecting                           & Executing                                \\
maneuvering                             & driving                               & positioning                           & optimal                             & steering,                                \\
                                        & environment.                          & and                                   & and safe paths                      & stopping,                                \\
                                        & Identifying safe                      & navigation                            & for navigation.                     & accelerating,                            \\
                                        & paths to navigate.                    & plans.                                &                                     & decelerating.                            \\
                                        & Identifying objects.                  &                                       &                                     &                                          \\ \hline
User interface                          & Monitoring state                      & Generating                            & Selecting the                       & Activating                               \\
                                        & of the robot                          & options for                           & means to                            & modes for                                \\
                                        & such as network                       & controlling                           & control                             & navigation,                              \\
                                        & connectivity,                         & the robot                             & the robot                           & and                                      \\
                                        & battery status,                       & and for                               & and to                              & communication.                           \\
                                        & required features                     & communication.                        & communicate.                        &                                          \\
                                        & on the interface.                     &                                       &                                     &                                          \\ \hline
\end{tabular}
\end{table*}

These functions were utilized in defining two LOA modes, to qualify the MRP system for navigation focused on the specific task of obstacle detection and avoidance (Table \ref{tab:2}). In this process, considerations were made regarding possible failures or emergencies. The role of handling such situations is referred to as a 'contingency role' which involves taking certain actions in the light of such events. Details of the roles in each of the LOA modes are described below. Some roles such as monitoring, analysis and selection of multimedia settings and communication are entirely under the operator's control in both modes.

\subsection{Assisted Teleoperation Mode (Low LOA)}
In the low LOA mode, the human observes the environment through a video stream. A map located on the side of the interface displays the robot's position and is continually updated as the robot moves. The human observes the environment to ensure that the robot does not collide with any object or person as it is directed towards a goal. Meanwhile, the robot scans the environment with its laser scanner. If obstacles are detected within a certain range, it decelerates to avoid collision or to attenuate the impact. This is done simultaneously and autonomously, without the need for the human to activate, manipulate or regulate the monitoring process during the task. This mode can be described as a guarded teleoperation mode.

The operator is further in charge of strategy generation and decision making, identifying possible trajectories to reach the chosen destination and managing various multimedia settings related to the social interaction. 
These actions are jointly executed by the human and the robot. The human takes the action of activating controls related to starting the robot, positioning and navigating the robot through the user interface, but the robot decelerates based on obstacle proximity as reported by its laser readings.

Additionally, it is primarily the human operator's responsibility to monitor the system's state through the user interface, to ensure that essential modules are functioning adequately. The robot only assists partially in monitoring the battery level, setting off a warning sound when it drops below a preset threshold.

\subsection{Autonomous Navigation Mode (high LOA)}
The robot is fully responsible for observing the environment to identify potential obstacles in its path, while the human operator only monitors the state of the robot through the user interface. The autonomous system identifies areas that are safe for navigation and within which the operator can select an appropriate goal location. The autonomous navigation, in turn, plots a viable path, if any can be found, towards the desired destination and subsequently follows it. As the robot moves through the environment, it continuously updates its local map to track dynamic obstacles which it avoids when and where possible. This differs from the assisted teleoperation mode, where the map is also continuously updated but the robot only decelerates to avoid or minimize collision with obstacles. As a special feature in this LOA mode, the operator can decide to switch between LOA modes or cancel an active goal, as deemed appropriate or necessary throughout the interaction. The rationale for allowing participants to switch to the low LOA was that, upon arriving at a location, teleoperation is better suited for fine-tuning, particularly for adjusting the orientation to face areas of interest. While it is possible to do so with autonomous navigation mode, it may not be quite as precise as desired.

\begin{table*}[!ht]
\centering
\caption{Summary of LOA Modes Implemented for the MRP}
\label{tab:2}
\begin{tabular}{lllllllll}
\hline
\textbf{LOA Mode}  & \textbf{Brief Description} & \multicolumn{4}{c}{\textbf{Function Allocation}}                    & \textbf{Cont.} & \textbf{Coord.} & \textbf{Switch} \\
 &                            & \textbf{Observe} & \textbf{Orient} & \textbf{Decide} & \textbf{Act} &                      &                      &  \\ \hline
Low LOA  & Human observes       & Human & Human & Human & Human & Human & Human & NA\\
(Assisted     & through robot,       & and   &       &       & and   &       &       & \\ Teleoperation) & plans and decides    & robot &       &       & robot &       &       &\\
     & alone but executes   &       &       &       &       &       &       &\\
     & with robot.          &       &       &       &       &       &       &\\ \hline
High LOA & Robot scans          & Robot & Robot & Robot & Robot & Robot & Robot & Human\\
(Autonomous     & with human but       & and   &       & and   & and   & and   & and   &\\
navigation) & plans navigation     & human &       & human & human & human & human &\\
     & alone. Human decides &       &       &       &       &       &       &\\
     & goal but robot       &       &       &       &       &       &       &\\
     & decides path.        &       &       &       &       &       &       &\\ \hline
\end{tabular}
\\ \textit{Cont=contingency roles for fallback tasks, Coord=coordinating responsibility demands\\ such as in path planning and obstacle avoidance, Switch=LOA switching option}
\end{table*}

\section{Methodology}
\label{sec:methodology}
\subsection{Overview}
\begin{figure}[ht]
\centering
\includegraphics[width=\linewidth]{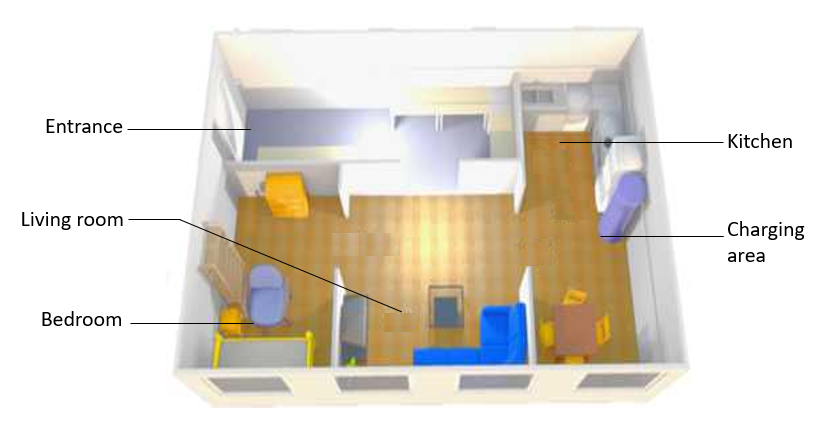}
\caption{A cross-section of the home-like environment used for the study.}
\label{fig:1}
\end{figure}

User studies were conducted at the Center for Applied Autonomous Sensor Systems (AASS) at Örebro University in Sweden. A telepresence robot was deployed in a home-like testbed environment (called 'PEIS-Home 2') with a living room, kitchen and bedroom (see Figure \ref{fig:1}). The walls separating the rooms are only approximately one meter high, which allows experimenters to keep a view over the entire space. In a separate room outside of view and hearing range of the robot, participants were seated in front of a standard computer screen with a mouse and keyboard to control the robot through a user interface. A social humanoid robot (Pepper robot)~\cite{pandey2018mass} was placed in the living room to represent the older adult living in the home (henceforth referred to as the \textit{resident robot}). Given its basic speech recognition function, we used it to evaluate how well the telepresence robot facilitates conversations, using a simple script which was provided for the participants. Pepper's height is comparable to that of a sitting adult and thus representative of a typical interaction scenario. The description of the MRP system, the LOA modes implemented on the user interface, the tasks and the experimental design are presented in the following subsections.

\subsection{The System}
\begin{figure}[ht]
    \centering
        \includegraphics[width=0.4\textwidth]{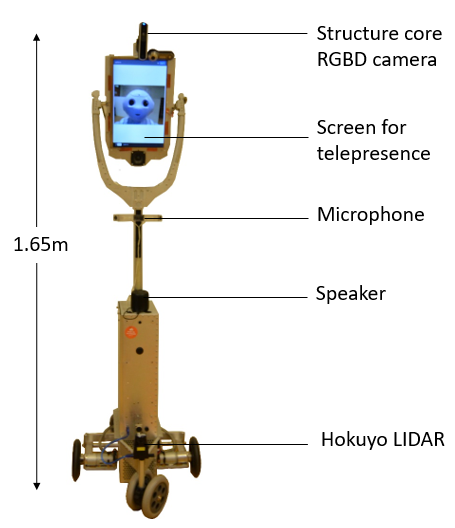}
    \caption{The modified Giraff telepresence robot.}
    \label{fig:2}
\end{figure}

The MRP system consists of a mobile robot platform with remote and local user interfaces developed to work through a server-client communication architecture utilizing WebRTC and rosbridge websocket. The user interfaces run within a standard web browser and are independent of the device's operating system or any specific software. This is particularly relevant in the case of the operator client, as it makes the robot more widely accessible and less prone to issues such as missing updates. More details on the robot platform and user interfaces are provided as follows:

\paragraph{The robot platform} The platform is an extensively retrofitted Giraff telepresence robot \cite{Orlandini2016} with a differential drive and screen with mechanical tilt function. Its height is approximately 1.65m and the footprint of the base 0.55 by 0.62m (see Figure \ref{fig:2}). The most significant modifications to the original hardware configuration include the removing of the plastic cladding, replacing the battery and adding a Hokuyo URG-04LX-UG01 2D laser scanner, as well as a wide-angle Structure Core RGBD camera. The latter was only used as an RGB webcam. The proprietary software was replaced with a custom-designed interface and back-end implementation based on the robot operating system (ROS). The autonomous navigation capability was realized via the ROS navigation stack, which provides the controller, local and global planners, as well as 2D obstacle detection using the laser range data.

\begin{figure*}[h]
    \centering
    \subfigure{\includegraphics[width=0.9\textwidth]{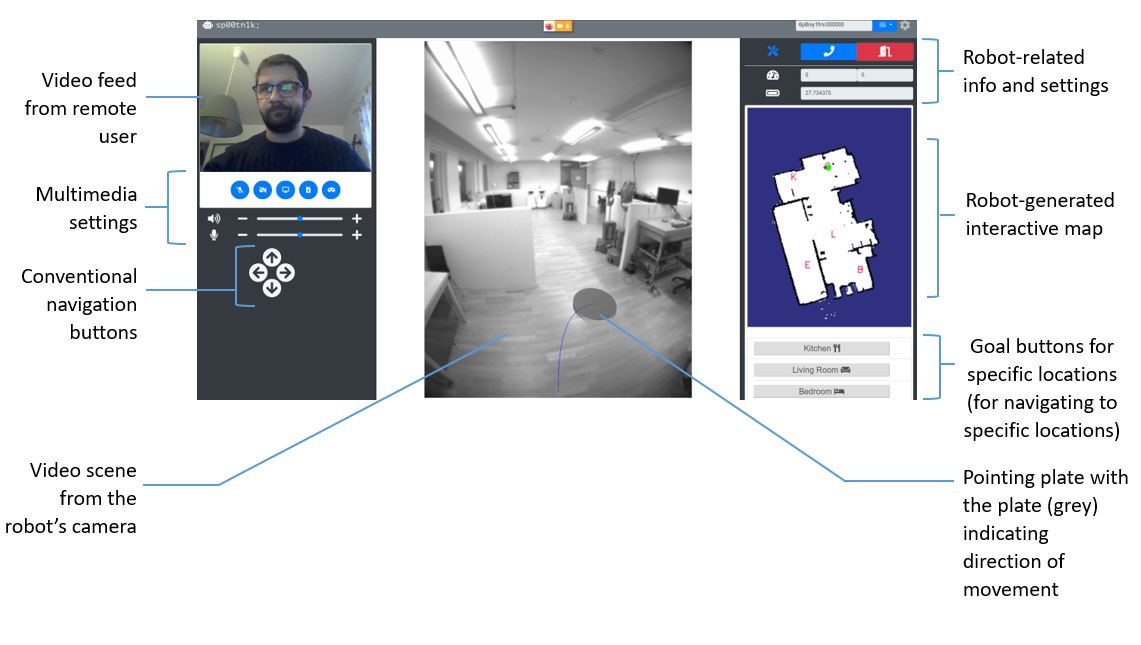}}\subfigure{\includegraphics[height=0.7in,width=0.1\textwidth]{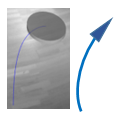}}
    \caption{The operator's user interface: The camera image from the robot with the graphical indicator for the manual navigation is at the center, the operator's own webcam, multimedia control buttons, and the conventional navigation buttons are located on the left of the screen; the interactive map and goal buttons are to the right. The same user interface was used for the high and low LOA mode but only specific features corresponding to the LOA in use were allowed for use in each experimental condition.}
    \label{fig:3}
\end{figure*}

\paragraph{User Interfaces} The remote operator user interface (see Figure~\ref{fig:3}) running on the client device is divided into three sections: a left, central and right panel. The left panel contains a small version of the operator's own video feed, as is common in videoconferencing, multimedia controls and the four conventional navigation control buttons. 
The central panel holds the video of the scene overlaid with a 'pointing plate' used as a trackball-like driving mechanism. More details on the use is provided in the next paragraph. The right panel includes a robot-generated reactive map of the apartment and 'goal buttons' to send the robot to specific locations within the home.

The two LOA modes are realized and accessed within this client user interface. In the low LOA mode, the first and main option for navigation control is operated via the pointing plate on the central panel of the interface. This pointing plate indicates the approximate position the robot would move towards, with a curved line originating from the center bottom to plot a rough trajectory towards that site (see Figure \ref{fig:3}). If the mouse is moved across the central part of the interface containing the video scene, the plate and tail follow the mouse on a plane projection representing the floor. By pressing and holding down the mouse button, users accelerate the robot; releasing the button causes it to slow down and eventually come to a stop. The higher up (and thus farther ahead of the robot) the cursor is on the image while the mouse button is pressed, the faster the robot moves forward. Likewise, the angular velocity increases as the cursor is dragged farther to the left and right edges. This type of interface was chosen because it allows for precision with respect to linear and angular velocity (as opposed to keyboard keys) while relying on the mouse as a commonplace input device. The other, auxiliary navigation control option for the low LOA mode was the conventional navigation control buttons on the left panel of the interface, i.e., arrows for left, right, forward, and backward in the X-Y plane. This option has been introduced largely as a fallback system for simple actions such as backing up or rotating in place.

In the high LOA mode the user selects the desired goal on the reactive top-down map on the right side of the interface. The map displays the environment used for the study, along with annotations of relevant locations corresponding to specific rooms. By clicking on any accessible location on the map, the robot is given the task to plan a suitable path and navigate towards the destination. Dragging the mouse in any direction before releasing the button determines the orientation the robot is going to rotate towards upon reaching the goal. The secondary navigation control option in the high LOA mode is given through the goal buttons, labeled with the name of the different locations in the environment (e.g. kitchen, living room, bedroom). The user can click on any of these buttons to send the robot to the desired location. Since this option only allows to move to a small set of discrete positions, it is not suited as a standalone function and rather intended to serve as a supporting feature. Nevertheless, both options rely on the same underlying mechanics (i.e., map coordinates and the ROS navigation stack). 

Switching between modes is an implicit action, as using one LOA mode deactivates the other. For example, clicking on the video image for manual teleoperation (low LOA mode) cancels any active goal of the autonomous navigation (high LOA mode). Similarly, manual teleoperation (low LOA mode) is effectively deactivated as soon as no input is given, while, as mentioned, clicking on the map activates high LOA mode.

In addition to the remote user interface, the local user interface is displayed on the robot's screen and includes the video stream for the local user to communicate with the remote operator.

\subsection{Tasks}
The tasks involved navigating the robot through the environment to specific locations while looking out for certain details in the surroundings. To assess participants' situation awareness, two items were placed on the floor in the doorway and the kitchen. The exact location of the items was randomized across the different trials. The robot always started at its charging station. An observation sheet was provided for the participants to take notes of their observations.
Below, we describe the action sequences in both tasks:

\paragraph{Low complexity task (LTask)}
This task involves a small set of actions and input sequence requirements from the user, contributing to low component and coordinative complexity (earlier defined), respectively. The sequence of actions is as follows:
1. Locate the resident robot in the environment and drive up to it so that its face is well visible in the interface,
2. Communicate with the robot (a short script for the communication was provided, consisting mostly of greetings which the resident robot had the ability to respond to),
3. Navigate the robot back to its charging area.

\paragraph{High complexity task (HTask)}
This task involves a more distinct set of actions, as the user has to carry out longer input sequences. This increases the component and coordinative complexity compared to the low complexity task. The sequence of actions is as follows:
1. Locate the charger of the resident robot,
2. Check if the door to the home is closed,
3. Check for any obstacles on the floor that could cause a fall (two specific objects were placed on the way to the home's entrance door and in the passage to the kitchen),
4. Check if the tap in the kitchen sink is running,
5. Navigate the robot to the charging area,
6. Check if the robot's charger is plugged in.

\subsection{Hypotheses}

Previous research revealed that performance is more likely to decline at higher robot autonomy with increasing task complexity~\cite{crandall2002characterizing}. Intuitively, it seems plausible to assume that longer and more complex tasks involve a higher potential for the occurrence of dynamic and unforeseen events which may require a degree of flexibility that (current) automation cannot provide.  We therefore propose that:
\\ \textit{H1: As task complexity increases, user performance will be higher in lower LOA (relative to higher LOA).}

High task complexity often involves a higher probability of performance declines, uncertainty and failure~\cite{onnasch2014human}. Previous research revealed that users appear to have more confidence in their own ability to handle decisions at such higher levels of complexity compared to automation-generated decisions~\cite{endsley1995out}. Increasing automation at high task complexity where more uncertainties can arise often shifts the workload towards monitoring, in an effort to ensure performance, as previously expressed by \cite{billings1991human,wiener1985cockpit}. In high-complexity situations, robot effectiveness is likely to decline if the robot is operating at higher autonomy, because there are higher chances for novel and dynamic conditions to arise for which the autonomous function was not prepared~\cite{goodrich2001experiments}. This calls for further investigation, particularly in the MRP scenario. We therefore define the second hypothesis to investigate this possibility as follows:
\\\textit{H2: As task complexity increases, workload will increase in higher LOA (relative to lower LOA).}

Assessments of SA at different levels of automation~\cite{endsley1995out} revealed that, as automation increases, users' comprehension of the different variables related to the current task declines. This trend appears self-evident, as higher LOA reduces the human involvement, thus allowing operators to direct their attention towards other tasks. This relates to an out-of-the-loop performance problem which has been widely documented as a potential negative consequence of higher LOA~\cite{billings1991human, moray1986monitoring, wickens2015engineering}. These works mostly examined tasks with professional operators in operational domains such as aviation. Some studies reveal minimal and in some cases no impact on SA, as reported in the meta-studies conducted in~\cite{onnasch2014human}. In the current MRP scenario, where caregivers, who are not robotic experts are expected to control the robot remotely, there are considerable differences with respect to system demands, user expectations and performance assessment. We therefore suggest that: 

\textit{H3: As task complexity increases, SA will be higher in lower LOA (relative to higher LOA).}

Many of the field studies conducted with MRP systems outlined the need for inclusion of autonomous features to improve the usability of the system and satisfaction ~\cite{Orlandini2016}. Efforts to introduce autonomy to MRP platforms to reduce mental demand on users have been reported to be promising~\cite{kiselev2015evaluation, kiselev2014combining}. The availability of more autonomous functions may increase users' willingness to use the system when facing increasingly complex tasks or higher workload. Therefore, we propose that: 
\\\textit{H4: As task complexity increases, the availability of LOA options will improve usability.}

\subsection{Experimental Design}
The study was conducted in a within-participants format, with every participant performing a total of four randomized trials corresponding to four different conditions. The conditions resulted from the 2x2 factorial combination of the independent variables, \emph{task complexity (low, high)}  and \emph{LOA mode (low, high)}. In the high LOA conditions, participants were allowed to switch to low LOA at any point in the task, as they preferred. This is due to the fact that, upon arriving at a location, teleoperation is often better suited for small adjustments to the position and orientation of the robot. These adjustments are frequently made when looking around and inspecting areas of interest. Therefore, we refer to this condition in the experiment as \emph{adjustable-high}, to better distinguish modes and experimental conditions.

The dependent variables were \emph{user performance}, \emph{perceived workload}, \emph{situation awareness}, and \emph{usability}, as detailed in section~\ref{subsec:measures}.

\subsection{Participants}
Twenty participants (7 female, 13 male) were recruited at Örebro University for the role of the caregiver (\emph{M=29, SD=6}). 
A majority (14) had a technical background (i.e. computer science, engineering), 6 of whom had worked with robots before. All reported that they use a computer daily, while most (16) use it at work consistently. 
Nine stated that they do not play any video games at all, while most of the others play games that may have a positive effect on their understanding of and performance with controlling the robot's interface (e.g., first person shooters, strategy games). All but two of the participants were affiliated with the university. 
All experiments were approved by the ethics review board at Örebro University.

\subsection{Procedure}
At the start of the experiment, after reading and signing the consent form, participants were 
asked to provide some background information regarding their age, gender, level of education, field of study/occupation as well as computer use and video gaming experience. 
Following this, they were briefed on the scenario, tasks and procedure. 
Before starting with the four main trials, users were introduced to the user interface and had the chance to practice with both operation modes until they felt sufficiently familiarized with the controls. This was defined as \emph{training to a basic use criterion} - to navigate the robot to a specified location and back. Each trial was followed by a questionnaire enquiring about the experience with the condition.
In between trials, while participants were occupied with the questionnaires, the experimenters would make subtle changes to the environment to reduce the learning effect. These changes concerned the locations or states of objects relevant to the tasks.
After completion of all four trials, participants answered a final questionnaire in which they rated their overall experience with the robot and tasks. It further afforded the opportunity to provide free input, feedback or remarks. 

\subsection{Measures}
\label{subsec:measures}

\subsubsection{Objective Measures}
For each participant and trial, performance was measured in terms of task completion time, the number of subtasks completed, and the number of obstacles missed (which is a subset of the subtask completed). The frequency and reason for switching of the LOA mode (switching from high to low LOA mode) during task execution was also taken into account. Additionally, the number of collisions while operating was recorded. Collisions were defined as any contact with walls, furniture or the resident robot, while assuming that all collisions are unintentional. Though not part of the objective performance assessment, it should be mentioned that the robot did occasionally collide in the high LOA mode, since its laser scanner only detects obstacles on a 2D plane. Few objects protruding at different heights were not registered. On other rare occasions the robot would rotate in place as a recovery behavior before continuing on to the destination. 

\subsubsection{Subjective Measures}
The post-trial questionnaires included a total of 19 questions from three questionnaires as detailed below. The questions were related to perceived workload (rated on a 5-point Likert scale), situation awareness (7-point Likert scale - according to the standard of the Situation Awareness Rating Technique (SART) \cite{taylor2017situational}, explained below) and usability (5-point Likert scale, see Table \ref{tab:3}). Perceived workload was assessed using the NASA-Task Load Index (NASA-TLX) questionnaire \cite{hart1988development}, with overall perceived workload rating computed from the different workload dimensions. It has previously been employed in the evaluation of MRP systems \cite{kiselev2012using}. Situation awareness was assessed using the 3D-SART version of the Situation Awareness Rating Technique \cite{taylor2017situational}, which measures complexity of interaction, focus of attention and information quantity. 
Subjective assessment concerning the system's usability was collected by means of the System Usability Scale (SUS) questionnaire \cite{brooke1996sus}. The final questionnaire included participants' assessments regarding the ease of use, as well as possible recommendations for how to develop the system further. 
Ease of use was evaluated with the Single Ease Questionnaire \cite{sauro201210}.

\renewcommand{\arraystretch}{1.2}
\begin{table*}[ht]
\caption{Self-report items from the post-trial questionnaire }
\label{tab:3}
\centering
\begin{tabular}{ p{.75\textwidth} }
\hline
\textbf{Perceived workload (scale 1-5) - NASA-TLX \cite{hart1988development}}\\
Mental Demand: How mentally demanding was the task? \\
Physical Demand: How physically demanding was the task? \\
Temporal Demand: How hurried or rushed was the pace of the task? \\
Performance: How successful were you in accomplishing what you were asked to do? \\
Effort: How hard did you have to work to accomplish your level of performance? \\
Frustration: How insecure, discouraged, irritated, stressed, and annoyed were you? \\
\hline
\textbf{System usability (scale 1-5) - SUS \cite{brooke1996sus}}\\
I think that I would like to use this system frequently. \\
I found the system unnecessarily complex. \\
I thought the system was easy to use. \\
I think that I would need the support of a technical person to be able to use this system. \\
I thought there was too much inconsistency in this system. \\
I would imagine that most people would learn to use this system very quickly. \\
I found the system very cumbersome to use. \\
I felt very confident using the system. \\
I needed to learn a lot of things before I could get going with this system. \\
\hline
\textbf{Situation awareness (scale 1-7) - 3D-SART \cite{taylor2017situational}} \\
Complexity of Interaction: Is it complex with many interrelated components (High) or is it simple and \\ straightforward (Low)? \\
Focus of Attention: Did you concentrate on many aspects of the interaction (High) or focus on only one (Low)? \\
Information Quantity: How much information have you gained about the environment the robot was navigating \\ in? \\
\hline
\end{tabular}
\end{table*}

\subsection{Data Analysis}
A generalized linear mixed model (GLMM) was applied to analyze the data with the LOA mode and task complexity as fixed modes, whereas the random effect was selected as the variances from the participants. The tests were designed as two-tailed with a significance level of 0.05.

\section{Results}
\label{sec:results}
Objective performance results are presented as a plot in Figure \ref{fig:4}, while more details on the interaction effects of LOA and task complexity on the subjective variables are provided in the following subsections:

\subsection{Objective Performance}
\begin{figure}[ht]
\centering
\includegraphics[width=\linewidth]{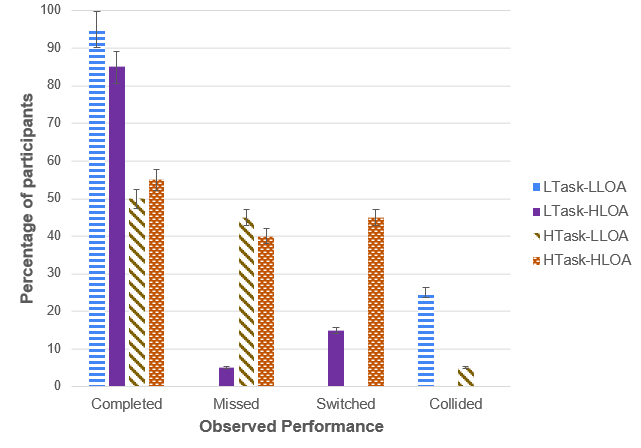}
\caption{Performance of the participants in the objective measures}
\label{fig:4}

\textit{LTask-LLOA = low LOA mode in low task complexity, LTask-HLOA = high LOA mode in low task complexity, HTask-LLOA = low LOA mode in high task complexity, HTask-HLOA = high LOA mode in high task complexity, Completed = Participants who completed all subtasks, Missed = Participants who missed obstacles, \\ Switched = Participants who switched to the low LOA mode while in high LOA mode, \\ Collided = Participants who collided with obstacles in the low LOA mode}
\end{figure}

\paragraph{Subtasks Completed}
In the low complexity task, all except one participant completed the subtasks using the low LOA mode and seventeen out of twenty participants completed the subtasks using the high LOA mode. In the high complexity task 10 of the 20 participants completed the subtasks using low LOA mode while 11 participants completed the subtasks using the high LOA mode. In the high LOA mode, 5 participants used the map option actively for navigation while about 4 participants stayed with the use of the conventional arrow key controls in the low LOA mode (see Figure~\ref{fig:4}). The 'fall risk' object which was placed on the floor in the robot's path was detected by 9 out of 20 participants in the high LOA mode and by 8 out of 20 participants in the low LOA mode in the high complexity task. One participant in the high LOA mode missed the resident robot in the low complexity task (Figure~\ref{fig:4}). 

\paragraph{Collisions} 
Three out of twenty participants had collisions with different objects in the environment (mainly with walls, a table and chairs) while navigating using the low LOA mode in the low complexity task and 4 participants had a collision in the high complexity task (Figure~\ref{fig:4}).

\paragraph{Switching of LOA} 
Twelve out of the twenty participants switched in the adjustable-high LOA condition to low LOA, on average of 2 times in a trial and for various reasons, as detailed in Table \ref{tab:4}. Three out of these twelve participants switched in the low complexity task, while 9 of the participants switched in the high complexity task. Six out of the twelve who switched to low LOA returned to the higher LOA after resolving the situation for which they made the initial switch (Figure~\ref{fig:4}). The remaining 6 of the 12 who switched to the low LOA kept switching between both modes until they completed the task. The reasons for this are also contained in the conditions in (Table~\ref{tab:4}). It should be mentioned that, if participants ended up executing more than half of the tasks with teleoperation in the adjustable-high LOA condition, it was counted as successful completion in low LOA mode. However, if the low LOA was only used for looking around and orienting the robot, it was considered as completed in the high LOA mode.

\begin{table}[ht]
\caption{Conditions observed for switching from the high LOA mode to the low LOA mode}
\label{tab:4}
\begin{tabular*}{\linewidth}{ll}
\hline
\textbf{Condition for Switching} & \textbf{Examples}                                                         \\ \hline
Automation transparency         & When they were not sure what the robot was \\ 
                                & doing while it was navigating autonomously \\
Unexpected events                & When the robot got stuck between obstacles \\
                                & and could not get out on its own
                                                                                             \\
Dissatisfaction                  & When the robot did not position itself or \\
with the automation              & accelerate as intended, participants switched   \\
of the robot                     & to the low LOA mode to adjust or fine-tune \\
                                 & the actions
                                                                                   \\
Error handling                   & If participants gave the wrong instruction to \\
                                 & the robot and wanted to correct the error 
                                 \\
Safety considerations            & Some of the participants stated that they \\
                                 & were concerned about the robot colliding  \\
                                 & with obstacles as it navigated on its own                                                \\
Automation malfunctions          & If the robot did not perform as expected in \\                                        &navigating to the desired location.          \\ \hline
\end{tabular*}
\end{table}

\subsection{Perceived Workload}

\paragraph{Mental demand} LOA mode and task complexity had a significant interaction effect on participants' mental demand (\emph{F(1,73)=1.589, p=0.009}). 
At low task complexity, the low LOA mode placed a higher mental demand (\emph{M=2.00, SD=1.05}) on the participants compared to the high LOA mode (\emph{M=1.53, SD=1.02}). 
In high task complexity, however, both low LOA mode (\emph{M=2.25, SD=0.97}) 
and high LOA mode (\emph{M=2.21, SD=1.18}) had an equally high mental demand. 
The LOA mode alone, however, did not have a significant effect on the mental demand (\emph{F(1,73)=1.910, p=0.171}), though at high task complexity (\emph{M=2.23, SD=1.06}) there was a significantly higher mental demand (\emph{F(1,73)=6.536, p=0.013}) compared to the low task complexity condition (\emph{M=1.76, SD=1.05}).

\paragraph{Physical demand} The interaction of the LOA mode and task complexity was significant with respect to the physical demand on the participants (\emph{F(1,73)=8.6, p=0.004}). 
Using the low LOA mode was similarly demanding in low task complexity (\emph{M=1.68, SD=1.00}) and high task complexity (\emph{M=1.65, SD=1.04}). 
However, while using the high LOA mode, participants reported a lower physical demand in low task complexity (\emph{M=1.16, SD=0.38}) compared to the high task complexity (\emph{M=1.79, SD=0.98}), where participants reported an increased physical demand. 
The task complexity as a main effect had a significant influence (\emph{F(1,73)=8.6, p=0.047}) 
on the physical demand in a pattern similar to the mental demand, in that the high complexity task was more physically demanding (\emph{M=1.72, SD=0.99}) than the low complexity task (\emph{M=1.42, SD=0.79}). 
Regardless of this, LOA mode as a main effect was not significant in terms of physical demand (\emph{F(1,73)=0.008, p=0.930}).

\paragraph{Temporal Demand} The interaction effect of the LOA mode and task complexity was not significant (\emph{F(1,73)=1.534, p=0.219}), neither was the LOA mode as a main effect (\emph{F(1,73)=0.383, p=0.538}). 
The task complexity was significant as a main effect (\emph{F(1,73)=6.136, p=0.016}). The high complexity task placed a higher temporal demand (\emph{M=2.05, SD=1.12}) on the participants compared to the low complexity task (\emph{M=1.63, SD=0.85}).

\paragraph{Operator Performance} 
Significant interaction was observed between LOA mode and task complexity (\emph{F(1,73)=4.376, p=0.04}) with respect to operator performance. In low task complexity, participants reported higher subjective performance (\emph{M=4.32, SD=1.00}) using the high LOA mode compared to the low LOA mode (\emph{M=3.74, SD=1.28}). At high task complexity, however, a reverse trend was observed -- the low LOA mode showed higher reported performance results (\emph{M=4.50, SD=0.76}) compared to the high LOA mode (\emph{M=4.16, SD=1.21}). 
The effect of LOA mode alone (\emph{F(1,73)=0.307, p=0.581}) or task complexity alone (\emph{F(1,73)=1.878, p=0.175}) did not have a significant effect on the reported performance.

\paragraph{Effort} The interaction between the LOA mode and task was not significant (\emph{F(1,73)=0.875, p=0.353}), neither was LOA mode alone as a main effect (\emph{F(1,73)=0.008, p=0.927}). As expected, in the high complexity task (\emph{M=2.41, SD=1.04}), participants reported more exertion of effort (\emph{F(1,73)=14.047, p$<$0.01}) than in the low complexity task (\emph{M=1.71, SD=0.90}).

\paragraph{Frustration} The interaction effect of LOA mode and task complexity was significant (\emph{F(1,73)=6.701, p=0.012}). In the low complexity task, participants reported higher frustration while using the low LOA mode (\emph{M=2.00, SD=1.16}) compared to the high LOA mode (\emph{M=1.47, SD=0.70}). The opposite was the case in the high complex task, where results revealed higher frustration in using the high LOA mode (\emph{M=2.32, SD=1.25}) compared to when using the low LOA mode (\emph{M=1.75, SD=1.12}). \\

The overall perceived workload rating for all workload dimensions confirmed that the difference in task complexity was significant (\emph{F(1,73)=17.51, p<0.01}). The high complexity task placed a higher demand (\emph{M=49.23, SD=12.90}) on the participants compared to the low complexity task (\emph{M=40.96, SD=10.31}). 

The correlation between participants' video gaming experience and the aggregated perceived workload scores, though slightly negative, was not found to be significant (\emph{r=-0.091, n=77, p=0.431}).

\subsection{Situation Awareness}

\paragraph{Complexity of Interaction}
The interaction of the LOA mode and task complexity was not significant (\emph{F(1,73) =2.378, p=0.127}).The LOA mode (\emph{F(1,73)=0.741, p=0.392}) and task complexity (\emph{F(1,73)=2.298, p=0.134}) also did not significantly influence the interaction complexity. However, participants perceived the interaction with the system to be more complex in the high complexity task when using the high LOA mode, which was not the case while using the low LOA mode.

\paragraph{Focus of Attention}
LOA mode and task complexity had an interaction effect on participants' focus of attention (\emph{F(1,68)=4.649, p=0.035}). While performing the low complexity task, participants concentrated their attention on more aspects of the interaction while using the low LOA mode (\emph{M=3.63, SD=1.74}) compared to the condition where they used the high LOA mode (\emph{M=2.79, SD=1.93}). This focus of attention was reversed in the high complex task, where participants concentrated on more aspects of the interaction in the high LOA (\emph{M=3.58, SD=1.92}) mode than in the low LOA mode (\emph{M=3.15, SD=1.84}).   

\paragraph{Information Quantity}
The perceived information quantity reported by the participants refers to the difference in information the participants gained about the environment for different LOAs at different task complexities. The interaction of LOA and task complexity did not have a significant effect on the perceived information quantity. (\emph{F(1,73)=0.179, p=0.674}). However, the pattern of information gained with respect to the LOA modes changed in the different task complexities, similar to the trend for focus of attention. In the low complexity task, participants reported that they gained more information about the environment when using the low LOA mode (\emph{M=5.58, SD=1.07}) compared to the high LOA mode (\emph{M=5.53, SD=1.12}). In the high complexity task, however, the opposite was the case -- participants reported to have gained more information while using the high LOA (\emph{M=5.42, SD=1.02}) relative to when using the low LOA mode (\emph{M=5.25, SD=1.37}).

\subsection{System Usability}
\label{subsec:usability}
An interaction effect was observed between the LOA mode and task complexity with respect to the confidence in the system while using it (\emph{F(1,73)=5.067, p=0.027}). In the low complexity task, participants reported higher confidence using the high LOA mode (\emph{M=4.53, SD=0.61}) compared to the low LOA mode (\emph{M=4.00, SD=0.94}), whereas in the high complexity task participants reported higher confidence using the assisted teleoperation mode (\emph{M=4.15, SD=0.81} vs. \emph{M=4.00, SD=1.00}).

The aspect of the SUS questionnaire evaluating the integration of the system's various functions was also significant (\emph{F(1,73)=4.013, p=0.049}) with respect to the LOA mode. Participants considered the system functions more integrated in the high LOA mode (\emph{M=4.24, SD=0.75}) compared to the low LOA mode (\emph{M=3.92, SD=0.87}).
The influence of the LOA mode and task complexity was not significant in the other aspects of the SUS questionnaire. The LOA mode, however, significantly influenced the aggregated usability score for all the participants (\emph{F(1,73)=4.174, p=0.045}). The usability scores were higher in the high LOA mode (\emph{M=59.00, SD=5.07}) compared to the low LOA mode (\emph{M=56.87, SD=6.01}). Although the mean scores were lower than the SUS-recommended 68\% \cite{brooke1996sus}, 85\% of the participants considered the system easy to use.

\section{Discussion}
\label{sec:discussion}
The presented study examines LOA implementation and use in the context of an MRP system operated through a user interface. The primary contribution was the development and evaluation of two LOA designs -- assisted teleoperation (low LOA) mode and autonomous high LOA) mode to support users in different task complexity situations. Some of the results should be taken with caution due to the relatively small number of participants; it is possible that more comprehensive user studies provide a more definitive picture on the relation between LOA and task complexity. 
In addition, we acknowledge that we did not evaluate dependency on user background factors, such as participants' age, gender and video gaming experience, which may have introduced bias into the results. 

\subsection{Impact of LOA in MRP systems}
The influence of the implemented LOA modes was observed only between the different task complexity levels. As the task complexity increased, the LOA which engaged participants more in managing the robot's functions yielded higher performance compared to the LOA with higher robot autonomy, in line with \textit{H1}. 

Previous studies \cite{wickens2010stages} and Onnasch \emph{et al.} \cite{onnasch2014human}, involving both expert and non-expert users revealed an overall improvement in performance with increasing automation for routine tasks. However, it was stated that in other situations involving tasks with more situational demands, critical decisions and action implementation, the performance declined with higher automation. This can be associated with those situations in our study in which the higher LOA (autonomous navigation mode) produced higher performance in the low complexity (and less demanding) task. In the high complexity task, which demanded more critical decisions and actions and in which more automation failures occurred, the lower LOA (assisted teleoperation mode) yielded higher performance. 

An increase in workload was observed as the task complexity increased while using the high LOA mode (in line with \textit{H2}). The reason for this might be connected with the frustration experienced by the participants in the higher task complexity when the automation failed or did not perform as expected (as seen in the frustration dimension of the NASA-TLX). In these situations, participants switched to lower automation (as was possible in the adjustable-high LOA condition), perhaps to facilitate easier handling of some of these challenges as noted in \cite{olatunji2019improving} where a lower LOA was found to better facilitate easier interaction. The switching however incurs some switch costs \cite{wylie2000task, kaber2018issues} which could have contributed to the frustration observed in the high LOA mode. This switch cost, may also have contributed to the reason why most of the participants did not switch when the task complexity was low. 

Consistent with previous findings that lower LOA tends to improve the SA of users \cite{onnasch2014human, endsley1995out}, in the low complexity task, the assisted teleoperation mode appeared to provide better situation awareness in terms of focus of attention and the information participants gained about the environment (in line with \textit{H3}). Moreover, participants seemed to miss fewer details about the environment, as seen in the objective measure assessing missed objects. This relation was reversed in the more complex task, which required a higher degree of awareness -- a higher level of automation produced higher SA, as participants seemed to be better able to detect the obstacles. This concurs with the view of \cite{kaber2018issues}, arguing that the outcome of LOA implementations may vary with different task demands and advocating for the characterization of these LOA models in different tasks, contexts and situations in order to collate the prevalent trends for model improvements. 

\subsection{Users' Attitudes Towards Automation}
\label{subsec:attitudes}
Most participants used the automated functions in the adjustable-high LOA condition as they were guided to use it. 
Their confidence in the automation appeared to increase with use, as observed in section~\ref{subsec:usability}. 
However, along with this rise in confidence, one of the observed behaviors was overreliance on the automation without recognizing its limitations. 

On the other hand, disuse of the automation was also observed, especially when the autonomous function failed, delayed or did not perform as expected. 
Some of the participants who switched to the low LOA mode mentioned that the robot did not provide sufficient information in such situations to enable them to take prompt actions. 
This might be attributable to a sense of responsibility for system outcome, which can positively affect error detection tendencies by the human operator \cite{kaber2018issues, kaber2013testing}.

Another observation was a form of satisficing behavior. Several participants did not exhaustively explore all control options, even though they could potentially yield better performance, e.g., the use of the map for autonomous navigation or the pointing plate (trackball-like interface) in the low LOA mode. Some simply continued using the conventional arrow control keys to navigate the robot since they were familiar with it and it presumably gave them a \textit{'good enough'} outcome. This behavior of accepting a readily identifiable operational solution that meets some minimum level of performance can mediate the use of automation, as noted by Kaber \emph{et al.}~\cite{kaber2013testing}. But it also has the tendency to lead to suboptimal solutions that could be detrimental to performance in general~\cite{kaber2018issues}.
Another group of participants was observed using the low LOA for looking around during adjustable-high LOA trials, i.e., autonomous navigation, to a specific location and then low LOA for orientation and inspection within short distances. This indicates the benefit of the adjustable LOA, which afforded the possibility of switching to a different LOA at will to better support specific task demands.

The need for exploration is, in part, the reason why we decided to allow switching between LOAs in the adjustable-high LOA condition. For example, when looking for a specific object, it could be a good strategy to set a goal in the approximate area of interest where the object is and then switch upon arrival. Once there, a participant would, if necessary, rotate in place via teleoperation to inspect the site more closely. In another case, the participant might see something interesting while an autonomous navigation goal is active. By clicking on the video (and thereby issuing a teleoperation command), the navigation goal is automatically canceled and, as before, the interesting object can be examined closer, either by moving towards it with teleoperation or by setting a new goal in front of it.
Although the inherent objective of the autonomous navigation, which is to reach the goal it is given in the most direct way possible (while avoiding collisions), may be considered as conflicting with the user's tasks to explore, we believe that it is this low-effort switching which negates the apparent mismatch of objectives, as the goal of the navigation stack in the autonomous function can be frequently updated and realigned to suit the purpose of exploration.

Presumptions were also made by the participants regarding the level of situation awareness that the camera afforded them. For example, in sequence 3 of the high complexity task, the participants were expected to inspect the whole floor area of the home environment for obstacles that could cause a fall. Some participants skipped the step, possibly because they presumed to have seen most of the floor area of the local environment while carrying out the preceding sequences. These instances were responsible for certain subtasks not being completed. The discrepancies observed between the reported situation awareness ratings in these trials and the objective results concerning missed objects highlight this point. It further points to the necessity of evaluating the behavioral patterns of users, which could include assumptions to be considered in automation design (as suggested in~\cite{kaber2018issues}).
 
The usability assessment outcome was lower than the recommended 68\% in SUS~\cite{brooke1996sus}. This may be due to the complexity of the user interface which the initial training did not completely overcome. The participants' comments reveal various areas of the design which could be amended to enhance the usability of the system. The comments are related to improved feedback from the interface, error handling, video quality of the robot's cameras and availability of zooming capabilities. Responses with regards to the ease of use and learnability of the LOA modes, however, revealed satisfactory outcomes. This highlights the potential for improved usability in complex tasks using the LOA options that are available (in line with \textit{H4}). 

Users seemed to adapt the control options within the LOA modes successfully, as suggested by their responses regarding satisfaction (detailed in section~\ref{subsec:usability} in the usability results). 
This aligned with the goal of including the control options in the design process to achieve seamless transitioning between control modes. Overall, participants' responses concerning the control options provided within each LOA mode in the different task complexity conditions revealed a relatively high degree of satisfaction.

\section{Conclusions and Future Work}
\label{sec:conclusions}

The user study yielded valuable insights into participants’ preferences and which characteristics of the operator interface related to LOA should be modified to enhance the user experience and performance. The evaluation of the impact of LOA and task complexity on performance, workload, situation awareness and usability is therefore a contribution that emerges as a building block for further studies which may include other forms and dimensions of complexity changes. It aids the design of automation which is based on empirical work rather than ad hoc designs relying on speculations and presumptions. 
Furthermore, the effect of real-time switching between LOA modes during task operations has been investigated in relation to task complexity, as a basis for further development of this strategy in LOA development and continuous adaptation to various task demands.

It is recommended that human operators be kept in the loop in all LOA modes. In the present context, this translates to using LOA modes that keep the users more involved in the task (such as the low LOA mode) to avoid overreliance, while improving detection of potential failures or conflicts. In addition, the higher LOA mode can be improved in the future to minimize the need for switching and fine-tuning. In situations when it is desired or required, however, switching should be effortless and seamless, incurring as little penalty to users' mental workload, situation awareness and trust as possible (following the results in section 4.4).

The users' attitudes towards the automation (section~\ref{subsec:attitudes}) can inform some recommendations for further LOA development in MRP systems, which include clarity of feedback from the robot to ensure that users remain aware of its actions at all times and in all LOA modes. 
Options for error handling are further recommended to be included in the LOA modes as part of the fallback mechanisms to employ when the robot fails (according to the comments in section~\ref{subsec:usability}). This encourages users to detect and resolve errors, provided the tools for resolution are made available. 

Ongoing work is focused on increased transparency of the high LOA mode and visual markers indicating close proximity to obstacles in the front and on the sides. We will further investigate effective measures to increase users' spatial awareness of the robot's immediate surroundings. This will require tests with different types of interfaces and cameras, along with their positioning on the platform. 
Moreover, future work should investigate additional interaction functions, e.g., zooming functions for the camera and extensive feedback from the robot.
By making these adjustments, we expect the performance and usability with the direct teleoperation mode to improve and task completion time to decrease.
In this way, we hope to provide a step forward in advancing MRP systems into markets and enabling them to become viable tools that add value to people's everyday lives.

\section*{Funding}
This work has received funding from the European Union’s Horizon 2020 research and innovation program under the Marie Skłodowska-Curie grant agreement No 721619 for the SOCRATES project. Partial support was provided by Ben-Gurion University of the Negev through the Helmsley Charitable Trust, the Agricultural, Biological and Cognitive Robotics Initiative, the Marcus Endowment Fund, the Rabbi W. Gunther Plaut Chair in Manufacturing Engineering and the George Shrut Chair in Human Performance Management.

\bibliographystyle{ieeetr}
\bibliography{references}

\end{document}